\documentclass{bmvc2k}

\usepackage{times}
\usepackage{epsfig}
\usepackage{graphicx}
\usepackage{amsmath}
\usepackage{amssymb}

\usepackage{bm}
\usepackage{multirow}
\usepackage{algorithm}
\usepackage{algpseudocode}
\usepackage{paralist}
\usepackage{color, colortbl}
\definecolor{Gray}{gray}{0.85}
\usepackage{balance}
\usepackage{wrapfig}

\title{Track Facial Points in Unconstrained Videos}

\addauthor{Xi Peng}{xipeng.cs@rutgers.edu}{1}
\addauthor{Qiong Hu}{qionghu.cs@rutgers.edu}{1}
\addauthor{Junzhou Huang}{jzhuang@uta.edu}{2}
\addauthor{Dimitris N. Metaxas}{dnm@cs.rutgers.edu}{1}

\addinstitution{
 Department of Computer Science\\
 Rutgers University\\
 New Jersey, USA
}
\addinstitution{
Department of Computer Science\\
The University of Texas at Arlington\\
Texas, USA
}

\runninghead{Peng, Xi}{Track Facial Points in Unconstrained Videos}

\begin{document}

\maketitle

\begin{abstract}
Tracking Facial Points in unconstrained videos is challenging due to the non-rigid deformation that changes over time. In this paper, we propose to exploit incremental learning for person-specific alignment in wild conditions. Our approach takes advantage of part-based representation and cascade regression for robust and efficient alignment on each frame. Unlike existing methods that usually rely on models trained offline,  we incrementally update the representation subspace and the cascade of regressors in a unified framework to achieve personalized modeling on the fly. To alleviate the drifting issue, the fitting results are evaluated using a deep neural network, where well-aligned faces are picked out to incrementally update the representation and fitting models. Both image and video datasets are employed to valid the proposed method. The results demonstrate the superior performance of our approach compared with existing approaches in terms of fitting accuracy and efficiency.
\end{abstract}

\section{Introduction}
Fitting facial landmarks on sequential images plays a fundamental role in many computer vision tasks, such as face recognition \cite{ParkhiBMVC15,TaigmanCVPR14}, expression analysis \cite{GuoTIP16,NicolaouIVC12}, and facial unit detection \cite{ValstarCVPR06,WuCVPR16}. It is a challenging task since the face undergoes drastic non-rigid deformations caused by extensive pose and expression variations, as well as unconstrained imaging conditions like illuminations changes and partial occlusions.

Despite the long history of research in rigid and non-rigid face tracking \cite{BlackCVPR95,PatrasFG04}, current efforts have mostly focused on face alignment on a single image \cite{CaoIJCV14,SaragihIJCV11,SunCVPR13,TrigeorgisCVPR16,TzimiropoulosCVPR15,XiongCVPR13,ZhangECCV14,ZhangTangECCV14,ZhuCVPR15,ZhuCVPR12}. They have shown great success with impressive results in standard benchmark datasets \cite{SagonasICCVW13,YangCoRR15}. However, when it comes to sequential images, many of them suffer from significant performance degradation especially in real-world scenarios under wild conditions \cite{ShenICCVW15}. They usually rely on models trained offline on still images and perform sequential alignment in a tracking-by-detection manner \cite{ChrysosICCVW15,ShenICCVW15,TangTIP12}. They lack the capability to capture neither the specifics of the tracked subject nor the imaging continuity in successive frames. To this end, personalized modeling rather than generic detection is preferred.

One rational way to achieve personalized modeling is to perform joint face alignment \cite{PengTPAMI10,SagonasCVPR14}, which takes the advantage of the shape and appearance consistency in a sequence to minimize fitting errors of all frames at the same time. However, these methods are restricted to offline tasks since they usually require all images are available before image congealing \cite{PengTPAMI10}. They also suffer from low-efficiency issue which severely impedes their performance on real-time or large-scale tasks \cite{PengICCV15}. 

To avoid these limitations, other approaches attempt to incrementally construct personalized models instead of joint alignment. They either adapt the holistic face representation using incremental subspace learning \cite{SungPRL09} or update the cascade mapping using online regression \cite{AsthanaCVPR14}. However, how to jointly update the both in a unified framework still remains an open question without investigation. Besides, former approaches often employ holistic models to facilitate the adaptation \cite{SungPRL09}, which is inferior to part-based models in challenging conditions \cite{SaragihIJCV11,ZhuCVPR12}. Moreover, many of them attempt to achieve personalized modeling without correction, which may inevitably result in model drifting.

In this paper, we further exploit person-specific modeling for sequential face alignment to address aforementioned issues. We first learn the part-based representation to model the facial shape and appearance respectively, as well as a cascade of nonlinear mappings from the facial appearance to the shape. The representation subspace and mapping parameters are then incrementally updated in a unified framework to achieve personalized modeling on the fly. In summary, our work makes the following {\bf contributions}: 

\begin{itemize}
\itemsep0em
\item We propose a novel approach for sequential face alignment. The person-specific modeling is investigated by incrementally learning the representation subspace and the cascade of regressors in a unified framework.
\item The proposed part-based representation together with the cascade regression guarantees robust alignment in unconstrained conditions. More importantly, they are crucial to efficiently construct personalized models for real-time or large-scale applications.
\item We propose to leverage deep neural networks for efficient and robust fitting evaluation. It significantly alleviates the drifting issue which would severely deteriorate learned models and inevitably lead to failure.
\end{itemize}

To fully evaluate the performance of our approach, we employed both image and video datasets in the experiments and compared our method with the state of the arts in terms of fitting accuracy and efficiency. We conducted detailed experimental analysis to validate each component of our approach. The results demonstrate that the proposed incremental learning can significantly improve the fitting accuracy with an affordable computational cost, especially in unconstrained videos with extensive variations.

\section{Related Work}
Face alignment in a single image has attracted intensive research interest for decades. Generally speaking, existing methods usually accomplish the task by learning a nonlinear mapping, which can be either regressors \cite{TzimiropoulosCVPR15,XiongCVPR13,ZhuCVPR15} or neural networks \cite{SunCVPR13,ZhangECCV14,ZhangTangECCV14}, from the facial representation, which is either holistic \cite{CootesTPAMI01,BlanzTPAMI03} or part-based \cite{SaragihIJCV11,ZhuCVPR12}, to landmark coordinates.

It has been proved that the part-based rather than the holistic representation is more robust to the extensive variations in unconstrained settings. For instance, Saragih {\em et al.} \cite{SaragihIJCV11} proposed the {\em regularized landmark mean-shift} (RLMS) to maximize the joint probability of the reconstructed shape based on a set of response maps extracted around each landmark using expectation maximization. Asthana et al. \cite{AsthanaCVPR13} proposed the {\em discriminative response map fitting} (DRMF) to learn boosted mappings from the joint response maps to shape parameters. Cao et al. \cite{CaoIJCV14} combined a two-level regression to achieve {\em explicit shape regression} (ESR) using shape-indexed features. Xiong et al. \cite{XiongCVPR13} proposed {\em supervised descent method} (SDM) to learn a sequence of descent directions using nonlinear least squares.

More recently, {\em deep neural networks} (DNNs) based methods have made significant progress towards systems that work in real-world scenarios \cite{TrigeorgisICML14,TrigeorgisNIPSW15}. For example, Sun {\em et al.} \cite{SunCVPR13} proposed to concatenate three-level convolutional neural networks to refine the fitting results from the initial estimation. Zhang \textit{et al.} \cite{ZhangECCV14} employed the similar idea of the coarse-to-fine framework but using auto-encoder networks instead of CNNs. Zhang \textit{et al.} \cite{ZhangTangECCV14} showed that learning face alignment together with other correlated tasks, such as identity recognition and pose estimation, can improve the landmark detection accuracy. 

The aforementioned methods have shown impressive results in standard benchmark datasets \cite{SagonasICCVW13}. However, they still suffer from limited performance in the sequential task as they completely rely on static models trained offline. To address this limitation, efforts of constructing person-specific models are made to improve the performance of sequential face alignment.

Some of them achieve person-specific modeling via joint face alignment. A representative example was proposed in \cite{SagonasCVPR14}, which used a clean face subspace trained offline to minimize fitting errors of all frames at the same time. However, these methods are usually limited to offline tasks due to their intensive computational costs. Others attempt to incrementally construct personalized models on the fly. For instance, Sung {\em et al.} \cite{SungPRL09} proposed to employ incremental principle component analysis to adapt the holistic AAMs to achieve personalized representation. Asthana {\em et al.} \cite{AsthanaCVPR14} further explored SDM in {\em incremental face alignment} (IFA) by simultaneously updating regressors in the cascade using incremental least squares. However, faithful personalized models can hardly be achieved without joint adaptation of the representation and fitting models in a unified framework. More importantly, blind model adaptation without correction would inevitably result in model drifting. How to effectively detect misalignment is still a challenging question that is seldom investigated. To address this issue, we propose a deep neural network for robust fitting evaluation to pick out well-aligned faces from misalignment, which are then used to incrementally update the representation subspace and fitting strategy for robust person-specific modeling on the fly.

\section{Our Approach}
In this paper, we propose a novel approach for face alignment in unconstrained videos. We first learn the {\em part-based representations} to model the facial shape and appearance respectively. The {\em discriminative fitting} is performed by learning a cascade of regressors that maps from the appearance representation to the shape parameters. Then personalized modeling is achieved by {\em incremental representation update} and {\em fitting adaptation in parallel}. Finally, we propose a deep fitting evaluation to alleviate the drifting issue. 

\subsection{Part-Based Representations}
Our goal is to jointly learn {\em the shape representation} and {\em appearance representation} using part-based models. Both representations should be compact and efficient to facilitate incremental person-specific modeling.

{\em The shape representation} is learned by firstly performing Procrustes analysis \cite{CootesCVIU95} on training images to obtain normalized facial shapes. Then we apply principle component analysis (PCA) to obtain the mean shape and eigenvectors $\{\mathbf{M}^s,\mathbf{V}^s\}$, where $s$ denotes shape. An instance shape can be modeled as $\mathbf{s}(\mathbf{p}) = \mathbf{M}^s + \mathbf{V}^s \mathbf{p}$, where $\mathbf{p}$ is the shape representation. 

{\em The appearance representation} is learned using local response maps \cite{SaragihIJCV11}. Given a image $\mathbf{I}$ and the shape representation $\mathbf{p}$, the local response map around the $l$-th landmark is $\mathbf{A}_l (\mathbf{p}, \mathbf{I}) = {1}/{(1 + exp(a_l \mathbf{\phi}(\mathbf{s}(\mathbf{p}),\mathbf{I}) + b_l))}$
, where $\{a_l,b_l\}_{l=1}^L$ are patch experts learned 6by cross-validation. $\mathbf{\phi}(\cdot)$ is the feature vector with a possible choice from SIFT, HOG, LBP, etc.

\begin{wrapfigure}{r}{0.5\textwidth}
  \begin{center}
    \includegraphics[width=0.46\textwidth]{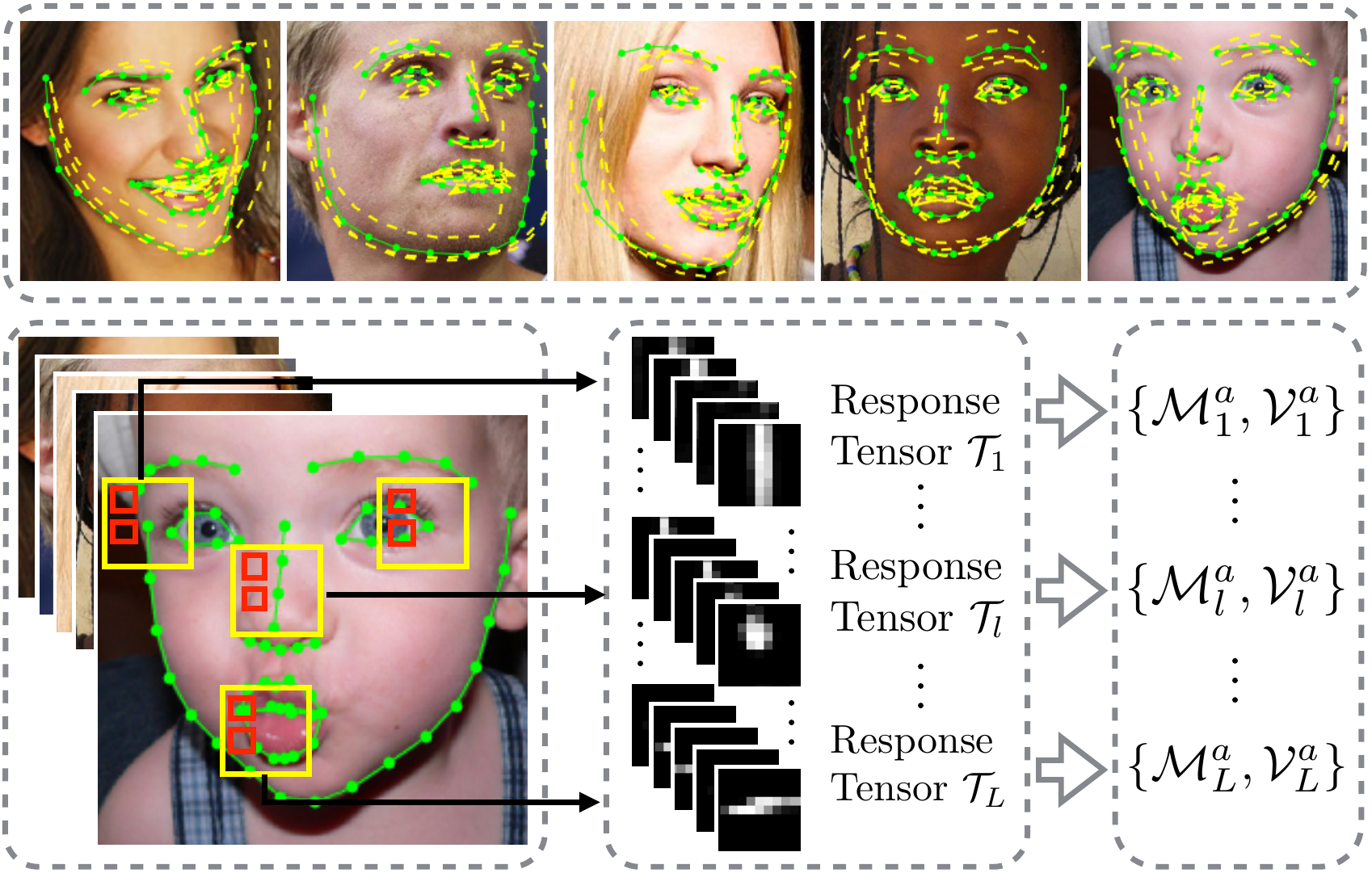}
  \end{center}
   \caption{Top: perturbations (yellow dash) are sampled around the ground-truth shape (green dash). Bottom: response maps (yellow box) of the same landmark are arranged as tensor to learn the appearance representation.} \label{fig:showperturb}
\end{wrapfigure}

To simulate the appearance variation and obtain more robust representation \cite{AsthanaCVPR13}, we sample perturbations $\{\Delta\mathbf{p}_{ij}\}$ around the ground-truth $\mathbf{p}_i^*$ as illustrated in Figure \ref{fig:showperturb}. The perturbed response maps are arranged as a tensor $\mathcal{T}_l=\{ \mathbf{A}_l(\mathbf{p}_i^*+\Delta \mathbf{p}_{ij}, \mathbf{I}_i) \}_{i,j}$, where $i$ and $j$ count images and perturbations respectively. Similar to the shape representation, we apply PCA on $\mathcal{T}_l$ to obtain the mean and eigenvectors $\{\mathbf{M}^a_l,\mathbf{V}^a_l\}$, where $a$ denotes appearance. The appearance representation of the $l$-th landmark can be calculated by fast projection $\mathbf{x}_l = (\mathbf{V}^a_l)^{-1}( \mathbf{A}_l(\mathbf{p},\mathbf{I}) - \mathbf{M}^a_l)$.

Now we can model the shape and appearance of an instance face using $\mathbf{p}$ and $\mathbf{x}(\mathbf{p},\mathbf{I}) = \begin{bmatrix} \mathbf{x}_1^T,\cdots;\mathbf{x}_L^T \end{bmatrix}^T $. The part-based representations are highly compact and efficient to compute. They are also robust to variations even for unseen images given the generative nature of parametric models \cite{PapaioannouTNNLS14,PengCVIU15}. These merits facilitate the incremental learning for person-specific modeling which will be explained soon in Section \ref{sec:iru}.

\subsection{Discriminative Fitting} \label{sec:df}
The goal is to learn a cascade of non-linear mappings from the appearance representation $\mathbf{x}(\mathbf{p},\mathbf{I})$ to the shape update $\Delta \mathbf{p}$. We refine the shape representation $\mathbf{p}$ from an initial guess $\mathbf{p}^0$ to the ground truth step by step:
\begin{equation} \label{eq:paramupdate}
\mathbf{p}^{k+1} = \mathbf{p}^{k} + \mathbf{x}(\mathbf{p}^{k},\mathbf{I}) \mathbf{R}^{k} + \mathbf{b}^{k},
\end{equation}
where $\{\mathbf{R}^k,\mathbf{b}^k\}$ is the regressor at step $k$ and $\mathbf{p}^*$ is the ground truth. Let $\Delta\mathbf{p}_{ij}^k = \mathbf{p}_i^\star - \mathbf{p}_{ij}^k$, the regressors can be computed by solving the least square problem \cite{XiongCVPR13}:
\begin{equation} \label{eq:mcsampling}
\underset{\mathbf{R}^k,\mathbf{b}^k}{\arg\min} \sum_{i=1}^{M} \sum_{j=1}^{N} || \Delta\mathbf{p}_{ij}^k - \mathbf{x}(\mathbf{p}_{ij}^k,\mathbf{I}_i) \mathbf{R}^k - \mathbf{b}^k || ^2.
\end{equation}
Let $\tilde{\mathbf{R}}^k = \begin{bmatrix} {\mathbf{R}^k}^T  {\mathbf{b}^k}^T \end{bmatrix}^T$ and $\tilde{\mathbf{x}} = \begin{bmatrix} {\mathbf{x}(\mathbf{p}^k,\mathbf{I}_i)}^T \mathbf{1} \end{bmatrix}^T$, the regressor can be computed with a closed-form solution
$\tilde{\mathbf{R}}^k = 
\begin{bmatrix}
\tilde{\mathbf{x}}^{T} \tilde{\mathbf{x}}+ \lambda \mathbf{I}
\end{bmatrix}
^{-1} \tilde{\mathbf{x}}^{T} \Delta\mathbf{p}^{k}
$.

Former approaches \cite{AsthanaCVPR13,CaoIJCV14} employed boosted regressors for discriminative fitting. However, it is difficult to perform incremental learning under the boosting framework due to the heavy computational load to update a large number of week regressors. In contrast, the cascade of regressors is easy to train, fast in test, and can be effectively adapted in parallel on the fly. We leave the details in Section \ref{sec:fap}.

\subsection{Incremental Representation Update} \label{sec:iru}
To achieve personalized representations of shape and appearance, our goal is to incrementally update the offline trained subspace $\{\mathbf{M}^s,\mathbf{V}^s\}$ and $\{ \mathbf{M}^a_l,\mathbf{V}^a_l \}_{l=1}^{L}$ in a unified framework. Suppose the offline model is trained on $m$ offline data $T_A$ with mean $M_A$ and eigenvectors $V_A$, where the SVD of $T_A$ is $T_A = U \Sigma V^T$. Given $n$ new online observations $T_B$ with mean $M_B$, our task is equivalent to efficiently compute the SVD of the concatenation $\begin{bmatrix} T_A~T_B \end{bmatrix} = U' {\Sigma}' {V'}^T$. 

It is infeasible to directly calculate the SVD as the entire offline training data need to be stored and reused online, which is extremely computationally expensive. Instead, we follow the {\em sequential Karhunen-Loeve} (SKL) algorithm \cite{LeveyTIP00,RossIJCV08} to formulate the concatenation as:
\begin{equation} \label{eq:ipcabig}
\begin{bmatrix}
U ~ E
\end{bmatrix} 
\begin{bmatrix}
\Sigma & U^T \hat{T}_B \\
\mathbf{0} & E (\hat{T}_B-UU^T\hat{T}_B)
\end{bmatrix}
\begin{bmatrix}
V^T & \mathbf{0} \\
\mathbf{0} & I
\end{bmatrix},
\end{equation}
where $ \hat{T}_B  =
\begin{bmatrix}
T_B ~ \sqrt{\frac{mn}{m+n}}(V_B - V_A)
\end{bmatrix}$, 
$E = orth(\hat{T}_B-UU^T\hat{T}_B)$.
Now we only need to perform SVD on the middle term instead of the entire concatenation:
\begin{equation} \label{eq:ipcatc}
T_C = \tilde{U}\tilde{\Sigma}\tilde{V}^T, \
T_C =
\begin{bmatrix}
\Sigma & U^T \hat{T}_B \\
\mathbf{0} & E (\hat{T}_B-UU^T\hat{T}_B)
\end{bmatrix}.
\end{equation}
By inserting $T_C$ back to Equation \ref{eq:ipcabig}, we have
$\begin{bmatrix}
T_A ~ T_B
\end{bmatrix}
=
\begin{pmatrix}
\begin{bmatrix}
U & E
\end{bmatrix}
\tilde{U}
\end{pmatrix}
\tilde{\Sigma}
\begin{pmatrix}
\tilde{V}^T
\begin{bmatrix}
V^T & \mathbf{0} \\
\mathbf{0} & I
\end{bmatrix}
\end{pmatrix}$.
The mean and eigenvectors can be instantly updated:
\begin{equation} \label{eq:ipca}
\begin{split}
M_{AB} & = \frac{m}{m+n} M_A + \frac{n}{m+n} M_B, \\
U' & = \begin{bmatrix} U~E \end{bmatrix} \tilde{U}, \ \Sigma'=\tilde{\Sigma}.
\end{split}
\end{equation}

Compared with the naive approach, the incremental subspace learning can significantly reduce the space complexity from $O(d(m+n))$ to $O(dn)$ and cut down the computational complexity from $O(d(m+n)^2)$ to $O(dn^2)$, where $m \gg n$ and $d$ denotes the length of a single observation. It guarantees efficient modeling of the personalized representations.

\subsection{Fitting Adaptation in Parallel} \label{sec:fap}
Once the shape and appearance representations are updated, we need to update the cascade of regressors instantly to catch up the online changes. However, adapting the cascade of regressors in a sequential order would be slow since $\tilde{\mathbf{R}}^k$ needs to be recomputed after $\tilde{\mathbf{R}}^{k-1}$. To address this issue, we follow \cite{AsthanaCVPR14} to decouple the dependence in the cascade by directly sample $\mathbf{p}^k$ from a norm distribution $\mathbf{p}^k \sim \mathcal{N}(\mathbf{p}^{\star}, \ \Lambda^k)$,
where $\Lambda^k$ is the shape variations learned offline. Once the cascade is flatten into independent mappings, all regressors can be simultaneously updated in parallel. 

During the offline training, we compute $\tilde{\mathbf{x}}_A$ and $\tilde{\mathbf{R}}_A$ following the definition given in Section \ref{sec:df}. During the online testing, we sample $\Delta \mathbf{p}_B$ based on the norm distribution and re-compute the new appearance representation $\tilde{\mathbf{x}}_B$. $\tilde{\mathbf{R}}_A$ can be adapted to $\tilde{\mathbf{R}}_{AB}$ by:
\begin{equation} \label{eq:illsw}
\begin{split}
\tilde{\mathbf{R}}_{AB} & = \tilde{\mathbf{R}}_A - P_{AB} \tilde{\mathbf{R}}_A + (P_A-P_{AB}P_A) (\tilde{\mathbf{x}}_B)^T \Delta \mathbf{p}_B,
\end{split}
\end{equation}
where $P_A = \begin{bmatrix} (\tilde{\mathbf{x}}_A)^T \tilde{\mathbf{R}}_A + \lambda I \end{bmatrix} ^{-1}$ , $P_{B} = \begin{bmatrix} \tilde{\mathbf{x}}_B P_A \tilde{\mathbf{x}}_B^T + I \end{bmatrix}^{-1}$, and $P_{AB} = P_A \tilde{\mathbf{x}}_{AB}^T P_{B} \tilde{\mathbf{x}}_B$.

Given the fact that $d \gg n$, the computational cost of the matrix inversion in Equation \ref{eq:illsw} is significantly reduced from $O(d^3)$ to $O(n^3)$ by decoupling regressors in the cascade. It is also highly memory-efficient since we can pre-compute $P_A$ offline and only a small number of online observations $\tilde{\mathbf{x}}_B$ need to be maintained for incremental adaptation.

\subsection{Deep Fitting Evaluation}
It is crucial to evaluate the fitting results since blind adaptation using erroneous fittings would inevitably result in model drifting. To address this issue, we leverage deep neural networks for robust fitting evaluation. Only well-fitted faces will be used to incrementally update the representation subspace and adapt the cascade of regressors for person-specific modeling

\begin{wrapfigure}{r}{0.5\textwidth}
  \begin{center}
    \includegraphics[width=0.45\textwidth]{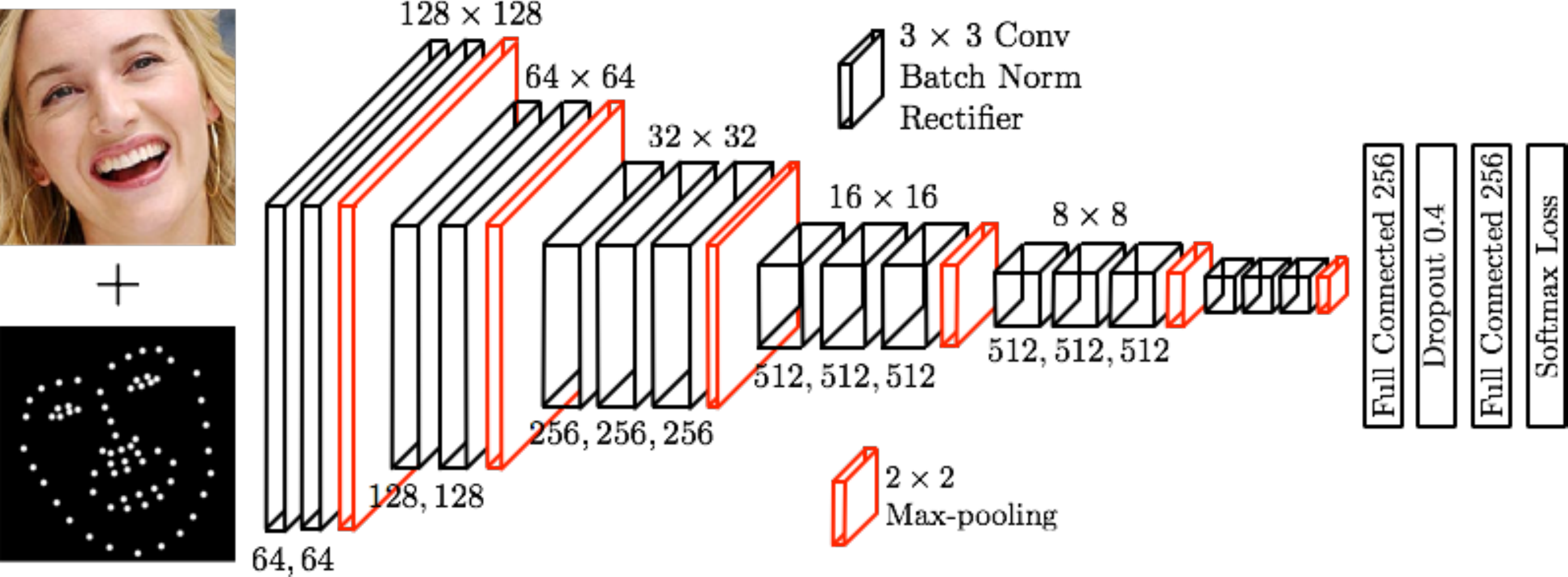}
  \end{center}
   \caption{The architecture of the fitting evaluation network. It takes the concatenation of the face image and landmark map as input and outputs a binary label to indicate correct or erroneous alignment.} \label{fig:dfe}
\end{wrapfigure}

Our goal is to learn a deep neural network that takes the fitting results as input and outputs a binary label to indicate correct or erroneous alignment. To connect the facial appearance and the fitted shape, a possible solution is to directly concatenate the vector of landmark coordinates to an intermediate fully connected layer \cite{WangWACV2015,WangIVC2016}. However, we experienced very limited performance using this design in our experiments. The reason is that the pixel-wise spatial information diminishes significantly after a series of max-pooling operations \cite{LongCoRR14}. The network can hardly learn the correct connection between the facial appearance and the landmark location. 

Instead, we propose to concatenate the facial image and the landmark map at the very beginning of the network as shown in Figure \ref{fig:dfe}. Each pixel in the landmark map is a binary value that marks the presence of the corresponding landmark. Our network is designed based on a variant of the VGG-16 networks \cite{SimonyanCoRR14} which has a reduced number of fully connected neurons. We can, therefore, initialize the training process from weights trained on large datasets for object classification. To fine-tune the network for our task, we construct a training set $\mathcal{U} = \{ (\mathbf{I}, \mathbf{S}); y \}$, where $y \in \{1,-1\}$. $\mathbf{I}$ is training images with landmark annotation. The landmark map $\mathbf{S}$ is generated using the ground-truth shape when $y=1$, or the perturbed shape when $y=-1$. We calculate cross-entropy loss for backpropagation.

The proposed deep fitting evaluation significantly outperforms former approach \cite{AsthanaCVPR14} that employs global and local handcrafted features for error detection, which will be discussed soon in Section \ref{sec:av}. It is also very efficient, which takes less than 10ms to process one image using a single K40 GPU accelerator.

\section{Experiments} \label{sec:experiment}
We first introduce the datasets used in our experiments as well as detailed settings. Then we perform algorithm validation and discussion to evaluate the proposed method in different aspects. Finially, we compare our approach with state of the arts in different datasets to demonstrate its superior performance.

\subsection{Datasets and Settings}
Both image and video datasets were used to conduct the experiments. The image datasets were mainly used to train the representation subspace and the cascade of regressors offline, while the video datasets were used to evaluate the performance of the proposed method. 

Four image datasets were used for offline training: {\bf (1) MultiPIE} \cite{GrossIVC10}, {\bf (2) LFPW} \cite{BelhumeurCVPR13}, {\bf (3) Helen} \cite{LeECCV12}, and {\bf (4) AFLW} \cite{Koestinger11}. From each of them, we collected 1300, 1035, 2330, and 4050 images of total 8715 images with 68-landmark annotations \cite{SagonasICCVW13}. 

Four video datasets were used for online testing: {\bf (1) FGNET} \cite{fgnet04}, {\bf (2) ASLV} \cite{NeidleLRECW14}, {\bf (3) 300-VW} \cite{ShenICCVW15}, and {\bf (4) YtbVW} \cite{PengICCV15}. From each of them, we collected 5, 10, 20, and 6 videos of more than 30,000 frames for the evaluation. These videos present unconstrained challenges, such as pose/expression variations, illumination changes, and partial occlusions.

We trained multi-view models based on different yaw angers \cite{PengFG15}: left $[-90^\circ,-30^\circ)$, frontal $[-30^\circ,30^\circ]$ and right $(30^\circ,90^\circ]$. All training images were registered to a reference 2D facial shape with an interocular distance of $50$ pixels to remove any 2D rigid movement. We employed HoG features to best balance the fitting accuracy and efficiency. The size of the patch expert and the local support window were set to $11 \times 11$ and $21 \times 21$ respectively. We sampled $10$ perturbations for each training image with the standard deviations of $\pm0.1$ for scaling, $\pm10^\circ$ for rotation, and $\pm10$ pixels for translation. Normalized Root Mean Square Error (Norm RMSE) was used in all experiments.

\subsection{Algorithm Validation and Discussion} \label{sec:av}
We conducted following experiments to validate the proposed approach in different aspects: person-specific modeling, joint adaptation, and deep fitting evaluation.

\begin{figure*}[t]
\minipage{\textwidth}
  \includegraphics[width=\linewidth]{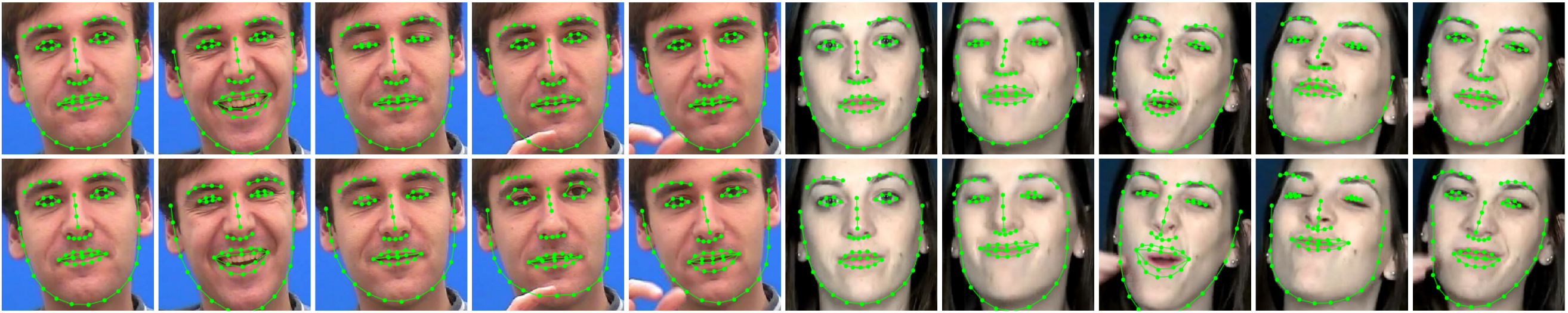}
\endminipage\hfill
\minipage{1\textwidth}
  \includegraphics[width=\linewidth]{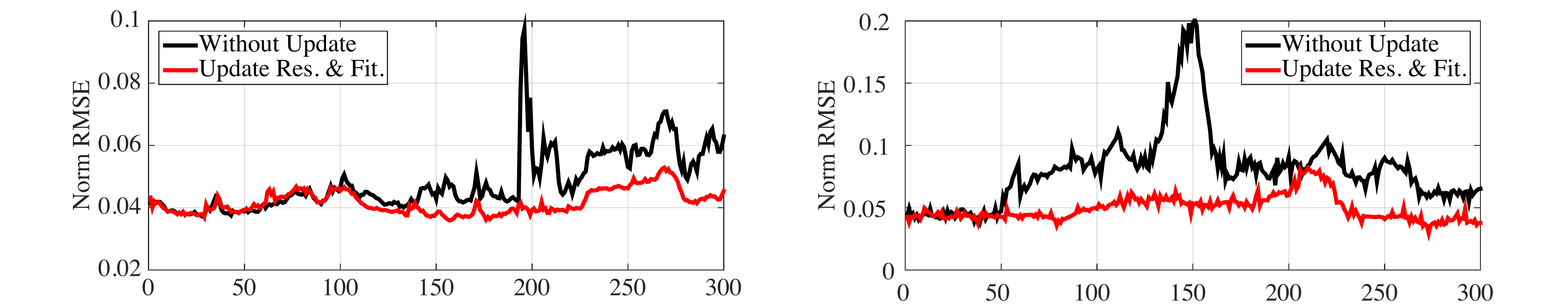}
\endminipage\hfill
\caption{Average fitting errors with and without the model adaptation on FGNET and ASLV.} \label{fig:updateornot_fig}
\end{figure*}

\textbf{Validation of person-specific modeling.} To investigate the contribution of the proposed personalized modeling, we first trained the representation and fitting models using MultiPIE and then collected two clips from FGNET and ASLV for testing. Each clip contains 300 frames with intensive pose and expression variations. The testing was performed under two different settings: (1) incrementally update the representation and fitting models, and (2) without any model adaptation. The frame-wise Norm RMSE in Figure \ref{fig:updateornot_fig} shows that both settings have comparable accuracy at the beginning. The online version outperforms the offline version once the model adaptation was performed. The superior performance becomes more significant when intensive variations and partial occlusions exist (around frame 200 of FGNET and frame 150 of ASLV).

\textbf{Validation of joint adaptation.} To investigate the joint adaptation of representation and fitting models, we first trained the offline models using all the training images and then carried out experiments on the full sets of FGNET and ASLV under three different settings: (1) update the representation model, (2) update the fitting model, and (3) update both models. The cumulative fitting errors are recorded in Table \ref{tab:updateornot}. The results indicate that only adapt the fitting strategy has better performance than only update the representation subspace. However, to achieve the best performance, it is necessary to jointly update both models in a unified framework. In this case, a more faithful personalized modeling can be expected by jointly update the representation subspace and adapt the fitting strategy.

\begin{table}[h]
\centering
\caption{Cumulative error distributions on FGNET and ASLV with different settings.} \label{tab:updateornot}
\begin{tabular}{ | c | c | c | c || c | c | c | }
\hline
\rowcolor{Gray}
 Update	 &	 $< 0.04$ & $< 0.06$ & $< 0.08$	& $< 0.04$ & $< 0.06$ & $< 0.08$	 \\
 \hline
 \hline
 Rep. & $78.2\%$ & $93.1\%$ &	$95.6\%$ & $62.9\%$ & $78.0\%$ &	$89.4\%$ 	 \\
 \hline
 Fit. & $84.0\%$ & $96.4\%$ &	$98.5\%$ &  $58.2\%$ & $73.3\%$ &	$91.2\%$ 	 \\
 \hline
 Rep. \& Fit. & $\textbf{91.8\%}$ & $\textbf{97.0\%}$ &	$\textbf{99.4\%}$ & $\textbf{68.7\%}$ & $\textbf{84.6\%}$ &	$\textbf{93.5\%}$	\\
 \hline
\end{tabular}
\end{table}

\textbf{Validation of deep fitting evaluation.} The online fitting evaluation is crucial in our incremental learning framework. Adaptation using erroneous fittings will drift the offline learned models and eventually lead to failure. To evaluate the performance the proposed deep fitting evaluation, we trained the network using image datasets and test the evaluation accuracy on videos. We sampled 5 perturbations for each image, where the ground-truth shape and perturbed shape were labeled as positive and negative respectively. Note that we used more negative rather than positive samples to train the network as misalignment detection is the major target here. Table \ref{tab:fiteval} shows the average misalignment detection accuracy in different video datasets. Our approach achieves around $90\%$ accuracy in general. It can robustly detect erroneous fittings in challenging conditions and pick out well-aligned faces for online adaptation, which can significantly alleviate the drifting issue.

\begin{table}[h]
\centering
\caption{Misalignment detection accuracy on different video datasets.} \label{tab:fiteval}
\begin{tabular}{| c | c | c | c | c |}
\hline
\rowcolor{Gray}
   & FGNET~\cite{fgnet04}  &   ASLV~\cite{NeidleLRECW14}   &  300-VW~\cite{ShenICCVW15} & YtbVW~\cite{PengICCV15} \\
\hline \hline
Accuracy		 	&	$94.4\%$ & $91.7\%$ & $85.3\%$ & $88.1\%$	\\
\hline
\end{tabular}
\end{table}

\subsection{Comparison with Previous Work}
We compare our approach with four approaches that reported state-of-arts performance: {\bf (1)} regularized landmark mean-shift (RLMS) \cite{SaragihIJCV11}, {\bf (2)} discriminative response map fitting (DRMF) \cite{AsthanaCVPR13}, {\bf (3)} incremental face alignment (IFA) \cite{AsthanaCVPR14}, and {\bf (4)} explicit shape regression face alignment (ESR) \cite{CaoIJCV14}. For a fair comparison, we tested these methods in a tracking-by-detection manner.


\textbf{Comparison of fitting accuracy.} We compared our approach with the four methods on different video datasets. The average fitting errors are compared in Table \ref{tab:cmp}. We have following observations. First, our approach has the lowest fitting errors and outperforms others with substantial margins, which demonstrates the superior performance of our approach in unconstrained videos. Second, compared with the performance on FGNET and ASLV, the advantage of our approach is more significant on 300-VW and YtbVW which present dynamic head movements, expression variations, illumination changes and partial occlusions. This result proves that the proposed person-specific alignment can better handle unconstrained data than other generic methods. Third, we also notice that ESR and IFA have better performance than RLMS and DRMF. A possible reason is that the explicit 2D shape used in ESR and IFA is more flexible than the constrained 3D shape used in RLMS and DRMF, which enables more accurate fittings when large pose and violent expression exist. However, they are still inferior to ours since they rely on offline models and lack the capability to capture the intensive online changes.

\begin{table}[h]
\centering
\caption{Comparison of the averate fitting errors of different methods on four datasets.} \label{tab:cmp}
\begin{tabular}{ | c | c | c | c | c |}
\hline
\rowcolor{Gray}
 &	 FGNET~\cite{fgnet04}  &   ASLV~\cite{NeidleLRECW14}   &  300-VW~\cite{ShenICCVW15} & YtbVW~\cite{PengICCV15}	 \\
 \hline
 \hline
 RLMS~\cite{SaragihIJCV11} & $4.11\%$ & $5.68\%$ &	$7.79\%$ & $7.19\%$  \\
 \hline
 DRMF~\cite{AsthanaCVPR13} & $3.75\%$ & $5.17\%$ &	$6.25\%$ & $6.03\%$ \\
 \hline
 IFA~\cite{AsthanaCVPR14} & $3.52\%$ & $4.54\%$ &	$5.71\%$ & $5.48\%$ \\
 \hline
 ESR~\cite{CaoIJCV14} & $3.49\%$ & $4.85\%$ &	$5.85\%$ & $5.61\%$	\\
 \hline
 OURS & $\textbf{3.36\%}$ & $\textbf{4.41\%}$ &	 $\textbf{5.38\%}$ & $\textbf{5.23\%}$	\\
 \hline
\end{tabular}
\end{table}

\textbf{Comparison of running time.} We compared the average running time per frame of different methods and report the results in Table \ref{tab:runtime}. For each method, the average speed was evaluated using the same 1000 frames. We tested the proposed method with either turning off or on the model adaptation. The results demonstrate that when the model adaptation is turned off, our approach is much more efficient than RLMS, and has comparable performance as DRMF and ESR. It slows down obviously when the incremental model adaptation is turned on. The reason is we apply the deep fitting evaluation at each frame, and perform the evaluation and model adaptation in a sequential order. The testing speed can be significantly improved with better implementation technique such as applying batch evaluation and model adaptation in parallel threads. We leave this as our future work.

\begin{table}[h]
\centering
\caption{Comparison of the average running time per frame of different methods.} \label{tab:runtime}
\begin{tabular}{| c | c | c | c | c |}
\hline
\rowcolor{Gray}
RLMS \cite{SaragihIJCV11}	&	DRMF~\cite{AsthanaCVPR13} &	ESR\cite{CaoIJCV14} & OURS (off)	& OURS (on)  \\
\hline \hline
116ms	 & 55ms & 89ms & 76ms & 218ms \\
\hline
\end{tabular}
\end{table}

\section{Conclusion}
In this paper, we propose a novel approach to track facial points in unconstrained videos. We investigate incremental learning to update the representation subspace and simultaneously adapt the cascade of regressors to achieve person-specific modeling. To address the drifting issue, we propose to leverage the deep neural network for robust fitting evaluation. Experiments on both image and video datasets have validated our approach in different aspects and demonstrated its superior performance compared with the state of the arts in terms of fitting accuracy and testing speed.

\bibliography{egbib}

\begin{thebibliography}{49}
\providecommand{\natexlab}[1]{#1}
\providecommand{\url}[1]{\texttt{#1}}
\expandafter\ifx\csname urlstyle\endcsname\relax
  \providecommand{\doi}[1]{doi: #1}\else
  \providecommand{\doi}{doi: \begingroup \urlstyle{rm}\Url}\fi

\bibitem[Asthana et~al.(2014)Asthana, Zafeiriou, Cheng, and
  Pantic]{AsthanaCVPR14}
A.~Asthana, S.~Zafeiriou, S.~Cheng, and M.~Pantic.
\newblock Incremental face alignment in the wild.
\newblock In \emph{IEEE Conference on Computer Vision and Pattern Recognition
  (CVPR)}, 2014.

\bibitem[Asthana et~al.(2013)Asthana, Zafeiriou, Cheng, and
  Pantic]{AsthanaCVPR13}
Akshay Asthana, Stefanos Zafeiriou, Shiyang Cheng, and Maja Pantic.
\newblock Robust discriminative response map fitting with constrained local
  models.
\newblock In \emph{IEEE Conference on Computer Vision and Pattern Recognition
  (CVPR)}, pages 3444--3451, 2013.

\bibitem[Belhumeur et~al.(2013)Belhumeur, Jacobs, Kriegman, and
  Kumar]{BelhumeurCVPR13}
Peter~N. Belhumeur, David~W. Jacobs, David~J. Kriegman, and Neeraj Kumar.
\newblock Localizing parts of faces using a consensus of exemplars.
\newblock In \emph{IEEE Transactions on Pattern Analysis and Machine
  Intelligence (PAMI)}, volume~35, pages 2930--2940, December 2013.

\bibitem[Black and Yacoob(1995)]{BlackCVPR95}
M.~Black and Y.~Yacoob.
\newblock Tracking and recognizing rigid and non-rigid facial motions using
  local parametric models of image motion.
\newblock In \emph{IEEE Conference on Computer Vision and Pattern Recognition
  (CVPR)}, pages 374--381, 1995.

\bibitem[Blanz and Vetter(2003)]{BlanzTPAMI03}
Volker Blanz and Thomas Vetter.
\newblock Face recognition based on fitting a 3d morphable model.
\newblock \emph{IEEE Transactions on Pattern Analysis and Machine Intelligence
  (PAMI)}, 25\penalty0 (9):\penalty0 1063--1074, Sep 2003.

\bibitem[Cao et~al.(2014)Cao, Wei, Wen, and Sun]{CaoIJCV14}
Xudong Cao, Yichen Wei, Fang Wen, and Jian Sun.
\newblock Face alignment by explicit shape regression.
\newblock \emph{International Journal of Computer Vision}, 107\penalty0
  (2):\penalty0 177--190, 2014.

\bibitem[Chrysos et~al.(2015)Chrysos, Antonakos, Zafeiriou, and
  Snape]{ChrysosICCVW15}
Grigoris~G. Chrysos, Epameinondas Antonakos, Stefanos Zafeiriou, and Patrick
  Snape.
\newblock Offline deformable face tracking in arbitrary videos.
\newblock In \emph{The IEEE International Conference on Computer Vision (ICCV)
  Workshops}, pages 954--962, 2015.

\bibitem[Cootes et~al.(1995)Cootes, Taylor, Cooper, and Graham]{CootesCVIU95}
T.F. Cootes, C.J. Taylor, D.H. Cooper, and J.~Graham.
\newblock Active shape models-their training and application.
\newblock \emph{Computer Vision and Image Understanding}, 61\penalty0
  (1):\penalty0 38 -- 59, 1995.
\newblock ISSN 1077-3142.

\bibitem[Cootes et~al.(2001)Cootes, Edwards, and Taylor]{CootesTPAMI01}
Timothy~F. Cootes, Gareth~J. Edwards, and Christopher~J. Taylor.
\newblock Active appearance models.
\newblock \emph{IEEE Trans. Pattern Anal. Mach. Intell.}, 23\penalty0
  (6):\penalty0 681--685, June 2001.
\newblock ISSN 0162-8828.

\bibitem[FGNet(2004)]{fgnet04}
FGNet.
\newblock Talking face video, 2004.
\newblock URL
  \url{http://www-prima.inrialpes.fr/FGnet/data/01-TalkingFace/talking_face.html}.

\bibitem[Gross et~al.(2010)Gross, Matthews, Cohn, Kanade, and
  Baker]{GrossIVC10}
Ralph Gross, Iain Matthews, Jeffrey Cohn, Takeo Kanade, and Simon Baker.
\newblock Multi-pie.
\newblock \emph{Image Vision Computing (IVC)}, 28\penalty0 (5):\penalty0
  807--813, May 2010.
\newblock ISSN 0262-8856.

\bibitem[Guo et~al.(2016)Guo, Zhao, and Pietikäinen]{GuoTIP16}
Yimo Guo, Guoying Zhao, and Matti Pietikäinen.
\newblock Dynamic facial expression recognition with atlas construction and
  sparse representation.
\newblock \emph{IEEE Transactions on Image Processing (TIP)}, 25\penalty0
  (5):\penalty0 1977--1992, 2016.

\bibitem[Koestinger et~al.(2011)Koestinger, Wohlhart, Roth, and
  Bischof]{Koestinger11}
Martin Koestinger, Paul Wohlhart, Peter~M. Roth, and Horst Bischof.
\newblock Annotated facial landmarks in the wild: A large-scale, real-world
  database for facial landmark localization.
\newblock In \emph{Workshop on Benchmarking Facial Image Analysis
  Technologies}, 2011.

\bibitem[Le et~al.(2012)Le, Brandt, Lin, Bourdev, and Huang]{LeECCV12}
Vuong Le, Jonathan Brandt, Zhe Lin, Lubomir Bourdev, and Thomas~S. Huang.
\newblock Interactive facial feature localization.
\newblock In \emph{European Conference on Computer Vision (ECCV)}, pages
  679--692, 2012.

\bibitem[Levey and Lindenbaum(2000)]{LeveyTIP00}
A~Levey and Michael Lindenbaum.
\newblock Sequential karhunen-loeve basis extraction and its application to
  images.
\newblock \emph{IEEE Transactions on Image Processing (TIP)}, 9\penalty0
  (8):\penalty0 1371--1374, 2000.

\bibitem[Long et~al.(2014)Long, Shelhamer, and Darrell]{LongCoRR14}
Jonathan Long, Evan Shelhamer, and Trevor Darrell.
\newblock Fully convolutional networks for semantic segmentation.
\newblock \emph{CoRR}, abs/1411.4038, 2014.

\bibitem[Neidle et~al.(2014)Neidle, Liu, Liu, Peng, Vogler, and
  Metaxas]{NeidleLRECW14}
Carol Neidle, Jingjing Liu, Bo~Liu, Xi~Peng, Christian Vogler, and Dimitris
  Metaxas.
\newblock Computer-based tracking, analysis, and visualization of
  linguistically significant nonmanual events in american sign language (asl).
\newblock \emph{LREC Workshop on the Representation and Processing of Sign
  Languages: Beyond the Manual Channel}, 2014.

\bibitem[Nicolaou et~al.(2012)Nicolaou, Gunes, and Pantic]{NicolaouIVC12}
Mihalis~A. Nicolaou, Hatice Gunes, and Maja Pantic.
\newblock Output-associative \{RVM\} regression for dimensional and continuous
  emotion prediction.
\newblock \emph{Image and Vision Computing (IVC)}, 30\penalty0 (3):\penalty0
  186 -- 196, 2012.
\newblock ISSN 0262-8856.
\newblock Best of Automatic Face and Gesture Recognition 2011.

\bibitem[Papaioannou and Zafeiriou(2014)]{PapaioannouTNNLS14}
A.~Papaioannou and S.~Zafeiriou.
\newblock Principal component analysis with complex kernel: The widely linear
  model.
\newblock \emph{IEEE Transactions on Neural Networks and Learning Systems},
  2014.

\bibitem[Parkhi et~al.(2015)Parkhi, Vedaldi, and Zisserman]{ParkhiBMVC15}
O.~M. Parkhi, A.~Vedaldi, and A.~Zisserman.
\newblock Deep face recognition.
\newblock In \emph{Proceedings of the British Machine Vision Conference
  (BMVC)}, 2015.

\bibitem[Patras and Pantic(2004)]{PatrasFG04}
I.~Patras and M.~Pantic.
\newblock Particle filtering with factorized likelihoodsfor tracking facial
  features.
\newblock In \emph{The IEEE International Conference on Automatic Face and
  Gesture Recognition (FG)}, pages 97--102, 2004.

\bibitem[Peng et~al.(2015{\natexlab{a}})Peng, Huang, Hu, Zhang, Elgammal, and
  Metaxas]{PengCVIU15}
Xi~Peng, Junzhou Huang, Qiong Hu, Shaoting Zhang, Ahmed Elgammal, and Dimitris
  Metaxas.
\newblock From circle to 3-sphere: Head pose estimation by instance
  parameterization.
\newblock \emph{Computer Vision and Image Understanding (CVIU)}, 136:\penalty0
  92--102, 2015{\natexlab{a}}.

\bibitem[Peng et~al.(2015{\natexlab{b}})Peng, Huang, Hu, Zhang, and
  Metaxas]{PengFG15}
Xi~Peng, Junzhou Huang, Qiong Hu, Shaoting Zhang, and Dimitris~N Metaxas.
\newblock Three-dimensional head pose estimation in-the-wild.
\newblock In \emph{The IEEE International Conference on Automatic Face and
  Gesture Recognition (FG)}, volume~1, pages 1--6, 2015{\natexlab{b}}.

\bibitem[Peng et~al.(2015{\natexlab{c}})Peng, Zhang, Yang, and
  Metaxas]{PengICCV15}
Xi~Peng, Shaoting Zhang, Yu~Yang, and Dimitris~N. Metaxas.
\newblock Piefa: Personalized incremental and ensemble face alignment.
\newblock In \emph{The IEEE International Conference on Computer Vision
  (ICCV)}, 2015{\natexlab{c}}.

\bibitem[Peng et~al.(2010)Peng, Ganesh, Wright, Xu, and Ma]{PengTPAMI10}
Y.~Peng, A.~Ganesh, J.~Wright, W.~Xu, and Y.~Ma.
\newblock {RASL: Robust Alignment by Sparse and Low-rank Decomposition for
  Linearly Correlated Images}.
\newblock \emph{IEEE Transactions on Pattern Analysis and Machine
  Intelligence}, July 2010.

\bibitem[Ross et~al.(2008)Ross, Lim, Lin, and Yang]{RossIJCV08}
David~A. Ross, Jongwoo Lim, Ruei-Sung Lin, and Ming-Hsuan Yang.
\newblock Incremental learning for robust visual tracking.
\newblock \emph{International Journal of Computer Vision (IJCV)}, 77\penalty0
  (1-3):\penalty0 125--141, May 2008.
\newblock ISSN 0920-5691.

\bibitem[Sagonas et~al.(2013)Sagonas, Tzimiropoulos, Zafeiriou, and
  Pantic]{SagonasICCVW13}
C.~Sagonas, G.~Tzimiropoulos, S.~Zafeiriou, and M.~Pantic.
\newblock 300 faces in-the-wild challenge: The first facial landmark
  localization challenge.
\newblock In \emph{The IEEE International Conference on Computer Vision (ICCV)
  Workshops}, 2013.

\bibitem[Sagonas et~al.(2014)Sagonas, Panagakis, Zafeiriou, and
  Pantic]{SagonasCVPR14}
C.~Sagonas, Y.~Panagakis, S.~Zafeiriou, and M.~Pantic.
\newblock Raps: Robust and efficient automatic construction of person-specific
  deformable models.
\newblock In \emph{IEEE Conference on Computer Vision and Pattern Recognition
  (CVPR)}, pages 1789--1796, June 2014.

\bibitem[Saragih et~al.(2011)Saragih, Lucey, and Cohn]{SaragihIJCV11}
Jason~M. Saragih, Simon Lucey, and Jeffrey~F. Cohn.
\newblock Deformable model fitting by regularized landmark mean-shift.
\newblock \emph{International Journal of Computer Vision (IJCV)}, 91\penalty0
  (2):\penalty0 200--215, January 2011.
\newblock ISSN 0920-5691.

\bibitem[Shen et~al.(2015)Shen, Zafeiriou, Chrysos, Kossaifi, Tzimiropoulos,
  and Pantic]{ShenICCVW15}
J.~Shen, S.~Zafeiriou, G.~Chrysos, J.~Kossaifi, G.~Tzimiropoulos, and
  M.~Pantic.
\newblock The first facial landmark tracking in-the-wild challenge: Benchmark
  and results.
\newblock In \emph{The IEEE International Conference on Computer Vision (ICCV)
  Workshops}, 2015.

\bibitem[Simonyan and Zisserman(2014)]{SimonyanCoRR14}
K.~Simonyan and A.~Zisserman.
\newblock Very deep convolutional networks for large-scale image recognition.
\newblock \emph{CoRR}, abs/1409.1556, 2014.

\bibitem[Sun et~al.(2013)Sun, Wang, and Tang]{SunCVPR13}
Yi~Sun, Xiaogang Wang, and Xiaoou Tang.
\newblock Deep convolutional network cascade for facial point detection.
\newblock In \emph{IEEE Conference on Computer Vision and Pattern Recognition
  (CVPR)}, pages 3476--3483, 2013.

\bibitem[Sung and Kim(2009)]{SungPRL09}
Jaewon Sung and Daijin Kim.
\newblock Adaptive active appearance model with incremental learning.
\newblock \emph{Pattern Recognition Letters (PRL)}, 30\penalty0 (4):\penalty0
  359 -- 367, 2009.

\bibitem[Taigman et~al.(2014)Taigman, Yang, Ranzato, and Wolf]{TaigmanCVPR14}
Yaniv Taigman, Ming Yang, Marc'Aurelio Ranzato, and Lior Wolf.
\newblock Deepface: Closing the gap to human-level performance in face
  verification.
\newblock In \emph{CVPR}, 2014.

\bibitem[Tang and Peng(2012)]{TangTIP12}
Ming Tang and Xi~Peng.
\newblock Robust tracking with discriminative ranking lists.
\newblock \emph{IEEE Transactions on Image Processing (TIP)}, 21\penalty0
  (7):\penalty0 3273--3281, 2012.

\bibitem[Trigeorgis et~al.(2016)Trigeorgis, Snape, Nicolaou, Antonakos, and
  Zafeiriou]{TrigeorgisCVPR16}
G.~Trigeorgis, P.~Snape, M.~A. Nicolaou, E.~Antonakos, and S.~Zafeiriou.
\newblock Mnemonic descent method: A recurrent process applied for end-to-end
  face alignment.
\newblock In \emph{IEEE International Conference on Computer Vision Pattern
  Recognition (CVPR)}, June 2016.

\bibitem[Trigeorgis et~al.(2014)Trigeorgis, Bousmalis, Zafeiriou, and
  Schuller]{TrigeorgisICML14}
George Trigeorgis, Konstantinos Bousmalis, Stefanos Zafeiriou, and Bjoern~W.
  Schuller.
\newblock {A Deep Semi-NMF Model for Learning Hidden Representations}.
\newblock In \emph{{International Conference on Machine Learning (ICML)}},
  2014.

\bibitem[Trigeorgis et~al.(2015)Trigeorgis, Nicolaou, Zafeiriou, and
  Schuller]{TrigeorgisNIPSW15}
George Trigeorgis, Mihalis Nicolaou, Stefanos Zafeiriou, and Bjoern~W.
  Schuller.
\newblock {Towards Deep Multimodal Alignment}.
\newblock In \emph{{NIPS Multimodal Machine Learning Workshop}}, 2015.

\bibitem[Tzimiropoulos(2015)]{TzimiropoulosCVPR15}
G.~Tzimiropoulos.
\newblock Project-out cascaded regression with an application to face
  alignment.
\newblock In \emph{IEEE Conference on Computer Vision and Pattern Recognition
  (CVPR)}, pages 3659--3667, 2015.

\bibitem[Valstar and Pantic(2006)]{ValstarCVPR06}
Michel Valstar and Maja Pantic.
\newblock Fully automatic facial action unit detection and temporal analysis.
\newblock In \emph{IEEE Conference on Computer Vision and Pattern Recognition
  Workshop (CVPRW)}, pages 149--, 2006.
\newblock ISBN 0-7695-2646-2.

\bibitem[Wang et~al.(2015)Wang, Guo, and Kambhamettu]{WangWACV2015}
Xiaolong Wang, Rui Guo, and Chandra Kambhamettu.
\newblock Deeply-learned feature for age estimation.
\newblock In \emph{Winter Conference on Applications of Computer Vision
  (WACV)}, pages 534--541. IEEE, 2015.

\bibitem[Wang et~al.(2016)Wang, Guo, Merler, Codella, Rohith, Smith, and
  Kambhamettu]{WangIVC2016}
Xiaolong Wang, Guodong Guo, Michele Merler, Noel~CF Codella, MV~Rohith, John~R
  Smith, and Chandra Kambhamettu.
\newblock Leveraging multiple cues for recognizing family photos.
\newblock \emph{Image and Vision Computing (IVC)}, 2016.

\bibitem[Wu and Ji(2016)]{WuCVPR16}
Yue Wu and Qiang Ji.
\newblock Constrained joint cascade regression framework for simultaneous
  facial action unit recognition and facial landmark detection.
\newblock In \emph{IEEE Conference on Computer Vision and Pattern Recognition
  (CVPR)}, 2016.

\bibitem[Xuehan-Xiong and {De la Torre}(2013)]{XiongCVPR13}
Xuehan-Xiong and Fernando {De la Torre}.
\newblock Supervised descent method and its application to face alignment.
\newblock In \emph{IEEE Conference on Computer Vision and Pattern Recognition
  (CVPR)}, 2013.

\bibitem[Yang et~al.(2015)Yang, Jia, Loy, and Robinson]{YangCoRR15}
Heng Yang, Xuhui Jia, Chen~Change Loy, and Peter Robinson.
\newblock An empirical study of recent face alignment methods.
\newblock \emph{CoRR}, abs/1511.05049, 2015.

\bibitem[Zhang et~al.(2014{\natexlab{a}})Zhang, Shan, Kan, and
  Chen]{ZhangECCV14}
Jie Zhang, Shiguang Shan, Meina Kan, and Xilin Chen.
\newblock Coarse-to-fine auto-encoder networks {(CFAN)} for real-time face
  alignment.
\newblock In \emph{European Conference on Computer Vision (ECCV)}, pages 1--16,
  2014{\natexlab{a}}.

\bibitem[Zhang et~al.(2014{\natexlab{b}})Zhang, Luo, Loy, and
  Tang]{ZhangTangECCV14}
Zhanpeng Zhang, Ping Luo, Chen~Change Loy, and Xiaoou Tang.
\newblock Facial landmark detection by deep multi-task learning.
\newblock In \emph{European Conference on Computer Vision (ECCV)}, pages
  94--108, 2014{\natexlab{b}}.

\bibitem[Zhu et~al.(2015)Zhu, Li, Loy, and Tang]{ZhuCVPR15}
Shizhan Zhu, Cheng Li, Chen~Change Loy, and Xiaoou Tang.
\newblock Face alignment by coarse-to-fine shape searching.
\newblock In \emph{IEEE Conference on Computer Vision and Pattern Recognition
  (CVPR)}, pages 4998--5006, 2015.

\bibitem[Zhu and Ramanan(2012)]{ZhuCVPR12}
Xiangxin Zhu and Deva Ramanan.
\newblock Face detection, pose estimation and landmark estimation in the wild.
\newblock In \emph{IEEE Conference on Computer Vision and Pattern Recognition
  (CVPR)}, 2012.

\end{thebibliography}
\end{document}